\documentclass[runningheads]{llncs}

\usepackage{graphicx}
\usepackage{amsmath,amssymb} 
\usepackage{bm}
\usepackage{color}
\usepackage{subcaption}
\usepackage[font=small,labelfont=bf]{caption}
\usepackage[width=122mm,left=12mm,paperwidth=146mm,height=193mm,top=12mm,paperheight=217mm]{geometry}

\usepackage[colorlinks]{hyperref}
\usepackage{algorithm}
\usepackage{algorithmic}

\usepackage{pdfpages}
\usepackage{pgfplots}
\usepackage{tikz}
\usepackage{epsfig}
\usepackage{array}
\usepackage{pifont}
\pgfplotsset{compat=newest}
\pgfplotsset{plot coordinates/math parser=false}
\pgfplotsset{tick label style = {font=\scriptsize}, every axis label = {font=\scriptsize}, legend style = {font=\scriptsize}, label style = {font=\scriptsize}}

\newlength\figureheight
\newlength\figurewidth
\usetikzlibrary{datavisualization}
\tikzset{every picture/.style={font issue={\fontsize{3.5}{2}}},font issue/.style={execute at begin picture={#1\selectfont}}}

\newcommand{\bx}{\boldsymbol{x}}

\newcommand{\sY}{\mathcal{Y}}

\newcommand{\sS}{\mathcal{S}}

\newcommand{\yhat}{\wh{y}}

\DeclareMathOperator*{\argmax}{argmax}
\DeclareMathOperator*{\argmin}{argmin}

\newcommand{\field}[1]{\mathbb{#1}}

\newcommand{\R}{\field{R}}

\newcommand{\wh}{\widehat}

\input{macronic}

\newcommand{\myparagraph}[1]{\noindent\paragraph{\textbf{#1}.}}

\usepackage{url}
\urldef{\mailr}\path|derosa@dis.uniroma1.it|
	\urldef{\mailt}\path|thomas.mensink@uva.nl|
	\urldef{\mailb}\path|caputo@dis.uniroma1.it|  

\begin{document}
\pagestyle{headings}
\mainmatter

\title{Online Open World Recognition}
\titlerunning{Online Open World Recognition- R. De Rosa, T. Mensink and B. Caputo}
\author{Rocco De Rosa  \and Thomas Mensink \and Barbara Caputo}
\institute{\mailr\\
	\mailt\\
	\mailb\\}

\maketitle

\begin{abstract}

As we enter into the big data age and an avalanche of images have become readily available, recognition systems face the need to move from close, lab settings where the number of classes and training data are fixed, to dynamic scenarios where the number of categories to be recognized grows continuously over time, as well as new data providing useful information to update the system. 
Recent attempts, like the open world recognition framework of Bendale et al \cite{bendale2015towards}, tried to inject dynamics into the system by 
incrementally adding new classes and detecting instances from unknown classes, while at the same time continuously updating the models for the known classes. 
In this paper we argue that to properly capture the intrinsic dynamic of open world recognition, it is necessary to add to these aspects (a) the incremental learning of the underlying metric, (b) the incremental estimate of confidence thresholds for the unknown classes, and (c) the use of local learning to precisely describe the space of classes.
We extend three existing metric learning algorithms towards these goals by using online metric learning.
Experimentally we validate our approach on two large-scale datasets in different learning scenarios. 
For all these scenarios our proposed methods outperform their non-online counterparts.
We conclude that local and online learning is important to capture the full dynamics of open world recognition. 
\end{abstract}

\section{Introduction}
 \vspace{-0.2cm}
\begin{figure}[t]
	\centering
	\includegraphics[width=.9\textwidth]{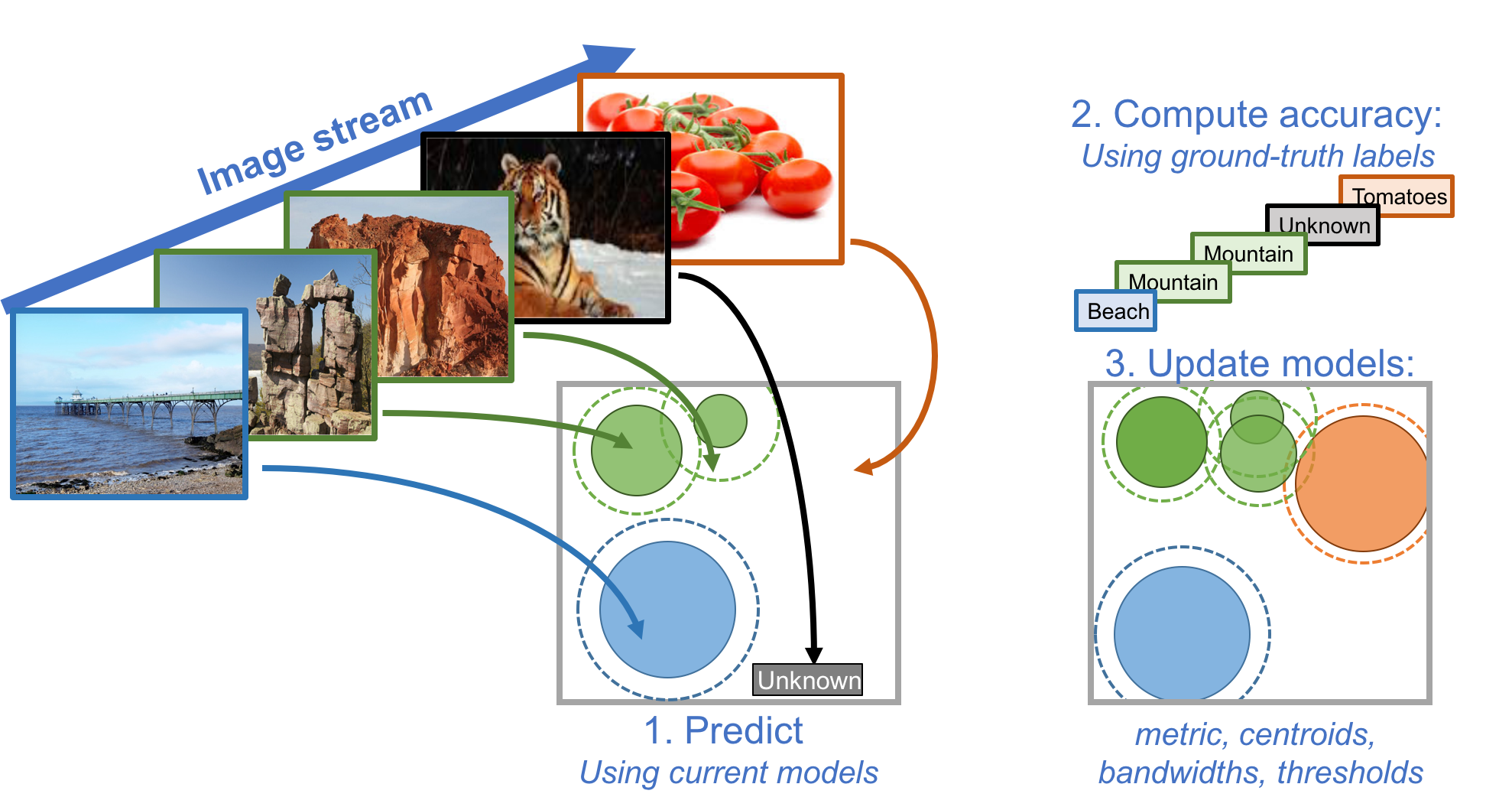}\vspace{-5mm}
	\caption{Our proposed online open world recognition workflow: as labeled data are presented continuously to the model, they are used to predict using the current classifiers, then to compute the accuracy, and finally to update the Mahalanobis metric, class centroids, bandwidths, and novelty thresholds incrementally. The resulting model is able to update continuously the internal representation of known classes, as well as detecting new one and adding them to the system on the fly.}
    \vspace{-0.5cm}
	\label{fig:learn_schema}
\end{figure}
 \vspace{-0.2cm}
The open world recognition framework has been introduced in 2015 by Bendale \emph{et al}~\cite{bendale2015towards}, as an attempt to move beyond the dominant classification methods assuming a static setting, where the number of training images is fixed as well as the number of classes that a model can handle. 
Its aim is to address the intrinsically dynamic nature of recognition in unconstrained settings, i.e. scenarios where it is not possible to predict a priori how many objects, and which, the system will have to recognize. 
This is true for robots equipped with cameras deployed in hospitals or  public spaces, or automatic tagging systems that have to deal with dynamically growing datasets, and so forth.

Open world recognition systems differ from standard, static visual classification algorithms in three key features: (a) their ability to incrementally update the model of the known categories as new data arrives; (b)  their ability to learn new categories, not seen initially during training, without the need to retrain the whole system from scratch, and (c) their ability to detect whether an incoming image depicts a known category, or if it is something new that needs to be learned. 
The requirement of adding new classes on the fly favours metric learning approaches (like $k$-nearest neighbours and nearest class mean classifiers) over SVMs~\cite{bendale2015towards}.
Several metric learning methods have been proposed so far, presenting some or all of these features~\cite{ncm_metric,ristin2015incremental,bendale2015towards}. 
Still, all these methods estimate the used metric, and the threshold for novelty detection, on an initial closed set of classes, and keep the metric and threshold fixed as the problem evolves. 
This conflicts with the very same definition of open world recognition, where the structure of the problem is progressively revealed as more data are observed, and the optimal parameters are likely to change over time. 

In this paper we argue that to properly model the dynamics of the challenging open world recognition scenario, it is necessary to learn online the \emph{metric} and the \emph{novelty threshold} as new instances and new classes arrive, rather than estimating them from an initial, closed set of classes as done so far~\cite{ncm_metric,ristin2015incremental,bendale2015towards}. 
This objective is similar as in online learning~\cite{shalev2011online} and stream mining~\cite{gama2013evaluating,derosa2015abacoc}.
Therefore we learn our classifiers online to incrementally update the model whenever new data is available, while at the same time being up-to-date for the predictions of both \texttt{known} (previously learned) classes and \texttt{unknown} classes (Figure \ref{fig:learn_schema}). 
Experimentally our incremental metric learning approaches demonstrate that continuously updating the metric as new data and new classes arrive leads to a 
better performance for both closed set accuracy and open set accuracy. 
Furthermore, we introduce a method to incremental learn the threshold for novelty detection, which uses the current internal confidences of the classifier for the $\texttt{known}$ classes.
This continuously tuning of the rejection threshold shows better performance as new classes are added to the classifier compared to a fixed threshold as previously used in~\cite{bendale2015towards}.
Our third contribution, is to introduce a non-linear local metric learning approach which adapts to the local complexity of the space with respect to the classes. 
Experimentally we show that this is especially beneficial in the open world recognition setting, since it is more flexible in modeling the border between known classes and unknown classes. 

Our findings are general, and applicable to a large class of algorithms. 
We demonstrate this by proposing online and incremental learning extensions of three non-parametric methods: 
(i) the Nearest Class Mean classifier (NCM) ~\cite{ncm_metric}, previously used for incremental adding novel classes in~\cite{ristin2015incremental};
(ii) the Nearest Non-Outlier classifier (NNO)~\cite{bendale2015towards}, which is an extension of NCM proposed for open world recognition;
(iii) the Nearest Ball Classifier (NBC)~\cite{derosa2015abacoc}, a local learning method incrementally adding balls (prototypes), and has been used in the streaming context before.
For all three algorithms, experiments show that the proposed extensions lead to a sizable advantage.
 \vspace{-0.2cm}
 \begin{algorithm}[t]                   
	\caption{Open World Online Learning Template}
	\label{alg:openworld}
	\begin{algorithmic}[1]                     
		\REQUIRE Initialise online performances, sample stream $(\bx_1,y_1),(\bx_2,y_2),\ldots$
		\FOR{$t=1,2,\ldots$}
		\STATE Receive sample $\bx_t$		
		\STATE Predict $\yhat_t \in \{\sY \cup \mathrm{\texttt{unknown}} \}$) using current models and  metric
		\STATE Receive true label $y_t$ 
        \STATE Output the online performances
		\STATE Update model and metric using $\bm x_t$, $y_t$ and the metric $W_{t}$.
		\ENDFOR
	\end{algorithmic}
\end{algorithm}
\vspace{-0.2cm}

\section{Related Work}
\label{sec:rw}

Our work is at the intersection of incremental and online learning, scalable learning, open set learning and open world recognition. In the following we will review previous work in the fields.

\myparagraph{Incremental Learning} There is a huge literature on incremental learning, such as various extensions of SVM~\cite{poggio2001incremental,yeh2008dynamic,pronobis2006more}. However, incremental SVMs suffer from several drawbacks, among which the most important is the extremely expensive update~\cite{laskov2006incremental}. There are some more efficient implementations~\cite{crammer2006online,shalev2011pegasos} but multi-class incremental learning does not permit the addition of new classes as well as other incremental classifiers~\cite{wang2010multi,li2010optimol}. Kuzborskij et al \cite{Kuzborskij_2013_CVPR} proposed a max-margin based approach for incremental learning of novel classes that exploited prior knowledge from previous classes, but the method had a conservative behavior, tending to privilege older classes with respect to the new one, performance-wise. 

\myparagraph{Scalable Learning} The goal of scalable systems is to achieve a good trade-off between prediction efficiency at test time and classification accuracy. Among these methods, tree-based approaches~\cite{marszalek2008constructing,liu2013probabilistic,deng2011fast} showed some success in addressing
scalability at test-time on large scale visual recognition challenges~\cite{everingham2010pascal,ilsvrc10}. Recently, these challenges have become dominated by deep learning methods~\cite{krizhevsky2012imagenet,szegedy14arxiv,simonyan2014very}. Again, the main drawback of these approaches is the need of a priori knowledge of the categories and of the availability of the whole training data during the learning phase. 

\myparagraph{Open Set Learning} Open set recognition considers the incompleteness about the knowledge of the world when learning a classifier, and the possible lack of knowledge of new classes during testing~\cite{li2005open,scheirer2013toward}. Scheirer et al.~\cite{scheirer2013toward} formulated the problem of open set recognition in a static one-vs-all setting  balancing open space risk and empirical error.
The setting was then extended~\cite{scheirer2014probability,jain2014multi} by introducing the compact abating probability model. This work offers robust methods to handle unseen classes. However, as it relies on the SVM decision scores, it does not scale. Fragoso et al.~\cite{fragoso2013evsac} proposed a scalable version  for modeling the matching scores, but they do not contextualized it in a  general recognition problem. A scalable incremental method on which we leverage on is the NCM classifier~\cite{ncm_metric}. 
Recently, NCM has been adapted for larger scale vision problems~\cite{ncm_metric,veenman2005nearest,veenman2005weighted,ristin2015incremental}, with the most recent approaches combining NCM with metric learning~\cite{ncm_metric} and with random forests~\cite{ristin2015incremental}. 
In contrast to the linear NCM classifier, the  nearest ball classifier (NBC)~\cite{derosa2015abacoc} is a non-linear local classifier. This incremental learning method adapts to the problem by adding new balls (prototypes). 
The NBC classifier has been used for classification in data streams~\cite{derosa2015abacoc} and action recognition in videos~\cite{de2014online}.
To the best of our knowledge the NBC has not been applied with metric learning nor for the open set recognition setting of this paper.

\myparagraph{Open World Recognition} Bendale and Boult further extended the notion of open set recognition to include incremental and scalable learning, leading to a more comprehensive problem that they called “open world recognition”~\cite{bendale2015towards}. To address it, the NCM algorithm was coupled with a module to limiting the open space risk for model combinations and transformed spaces, resulting in a new model, the nearest-non outlier (NNO) described in \sect{owc}.

\section{Online Open World Recognition}
In this section we introduce the online and incremental metric learning extension to three recent non-parametric classifiers. These classifiers will then be used within our open world online learning template, described in \alg{openworld}, to predict the label of each incoming sample.

\subsection{Closed Set Multi-Class Prediction}

For the closed-set multi-class prediction we focus on Nearest Class Mean classifiers (NCM).
They assign an instance to the class $y\in\sY$, where  $\sY = \{1,\dots,C\}$ is the set of possible classes, with the nearest mean vector $\bm\mu_y$~\cite{webb02spr}. 
Following~\cite{ncm_metric,bendale2015towards}, we use a multi-class probabilistic interpretation of NCM, and define the probability for class $y$ as: 
\begin{align}
    p(y|\bm x) 			&= \frac{\exp \big(-\!\tfrac{1}{2} \ d_W(\bm x,\bm\mu_y)\big)}{\sum_{y'\in\sY} \big(-\!\tfrac{1}{2} \ \exp d_W(\bm x,\bm\mu_{y'}) \big)}\label{eq:ncm_prob},
\end{align}
this is a soft-max function over the instance-to-class (squared) low-rank Mahalanobis distances $d_W(\cdot,\cdot)$, parameterized by $W$:
\begin{align}
d_W(\bm x,\bm \mu) 	&= (\bm x - \bm\mu)^{\top} W^{\top} W (\bm x - \bm\mu),
\end{align}
where $\bm x$ and $\bm\mu$ are $d$-dimensional vectors and $W \in \mathbb{R}^{m\times d}$, with $m \le d$ acting as regularizer\footnote{Also related in literature as the intrinsic dimension of the space.}, which improves computational efficiency.
Metric learning is used to find the best low-rank Mahalanobis distance, by optimizing the log-likelihood for correct classification over a training data-set:
\begin{align}
\logl &= \frac{1}{N} \sum_i \log( p(y_i|\bm x_i)).
\end{align}
Once a metric $W$ has been learned on a large set of classes, the obtained distance function has been shown to generalize for classifying novel classes~\cite{ncm_metric}. 
However, \textit{all} novel instances are used to set the class mean vectors, and the metric $W$ is not updated for those novel classes. 
In contrast, below we describe a method which learns incrementally both the class means and the metric $W$.

\myparagraph{Incremental learning}
In our scenario, the number of classes is unknown upfront and may change over time, therefore we learn the metric in an online fashion.
Given an example $(\bm x_t,y_t)\in\R^d\times\sY$, we update the NCM classifier as follows:
\begin{align}	
	\bm \mu^{t+1}_{y_t} &= \left(1-\frac{1}{n(y_t)}\right) \ \bm \mu^{t}_{y_t} + \frac{1}{n(y_t)} \bm x_t, \quad\quad\textrm{and}\label{eq:inc_ncm_m}\\
    W^{t+1} &= (1-\gamma) \ W^{t} + \gamma \ \gradient{W^t}{\log p(y_t|\bm x_t)},\label{eq:inc_ncm_w}
\end{align}
where $n(y_t)$ denotes the number of instances to class $y_t$ (including the example of time step t) and $\gamma$ is a fixed learning rate.
Note that the initial mean of a class $y$ always equals to the first observation $\bm x_t$ of that class: $\bm \mu^{t+1}_y = \bm x_t$.
The gradient of $W^t$ \wrt the model is given by:
\begin{align}
    \gradient{W^{t}}{\log p(y_t|\bm x_t)} &= \sum_{y\in\sY} \left(p(y_t|\bm x_t) - \ind{y_t=y} \right) \ W^{t} \  \left(\bm \mu^{t+1}_y- \bm x_t\right) \left(\bm \mu^{t+1}_y- \bm x_t\right)^{\top},\label{eq:ncm_gradient}
\end{align}
where we use Iverson brackets $\ind{\cdot}$ to denote the indicator function. The matrix $W^1$ is initialized by the truncated identity matrix, so it resembles the Euclidean distance. 
The metric update could be seen as a single step of stochastic gradient descent used in the large-scale closed set setting~\cite{ncm_metric}. 

The NCM classifier is not designed to predict whether an instance is from an unknown class or from the set of known classes. To accommodate for novelty prediction, we next describe the Nearest Non-Outlier algorithm for the open world classification scenario.

\begin{figure}[t]
	\centering
	\includegraphics[width=.8\linewidth]{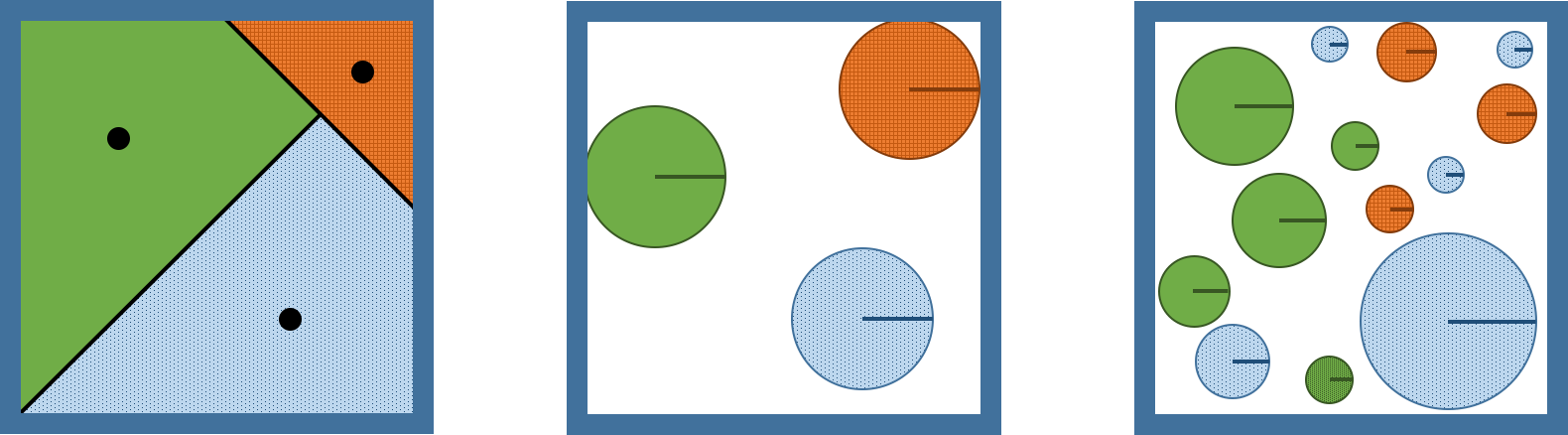}
    \caption{Illustration of different learning settings. In closed-set recognition (\emph{left}) the whole space is assigned to a specific class, while in open recognition (\emph{middle} and \emph{right}) classes have clear boundaries. Local learning (\emph{right}) allows for more flexible class boundaries which are useful in the open world recognition setting.}
    \label{fig:nno-illustration}
    \vspace{-0.5cm}
\end{figure}

\subsection{Open World Classification}
\label{sec:owc}

The Nearest-Non Outlier method is an extension of NCM for the open world scenario~\cite{bendale2015towards}, where NCM  is adjusted to define class boundaries, and instances beyond the class boundaries are assigned to the unknown class (\fig{nno-illustration}). 
Instead of using multi-class probability as defined in~\eq{ncm_prob}, in NNO the confidence score for class $y$ is given by:
\begin{align}
	s_y(\bm x,\tau) = Z_{\tau} \left(1 - \frac{1}{\tau} d_W(\bm x,\bm \mu_y)\right),\label{eq:nno_conf}
\end{align}
where $\tau$ is a threshold value to determine a ball around each class mean, and $Z_{\tau} = (\Gamma(\frac{m}{2}+1))/(\pi^{\frac{m}{2}} \tau^m)$ is a normalization factor to assure that $s_y$ integrates to 1 on the domain $s_y(\cdot) > 0$ (using the standard gamma function $\Gamma$).
An example $\bm x$ is rejected for class $y$ when $s_y(\bm x)\leq 0$, and assigned to the \texttt{unknown} class when it is rejected by all classes. 
In~\cite{bendale2015towards} the metric $W$ of NNO is learned offline on an initial set of known classes. 

\myparagraph{Incremental learning and rejection} We extend NNO to allow for incremental learning of the metric $W$ and automatically tuning of the class-rejection threshold $\tau$. 
We formulate the prediction confidence similarly to the RBF-Kernel:

\begin{align}
	C_y(\bm x_t,\theta^t) = \exp\left( - \frac{1}{2\theta^t} \ d_{W^t}(\bm x_t,\bm \mu^t_y)\right). \label{eq:onno_rbf}
\end{align}
This assigns a confidence value between $[0,1]$ to the sample $\bm x_t$ at time step $t$ for class $y$, using the current metric $W^t$.
The advantage of this RBF formulation is that the function is strictly bounded.
Using \eq{onno_rbf} also reduces the open space risk as defined in~\cite{bendale2015towards}, since it obeys to the abating property~\cite{scheirer2014probability}, given that the function value decreases in areas away from the observed training data. 
The bandwidth parameter $\theta^t$ is learned incrementally, using the expected value of distances to all class means (initialized with $\theta^1=1$):
\begin{align}
	\theta^{t+1} &= 
    (1-\tfrac{1}{t}) \ \theta^{t} + \frac{1}{t} \sum_{y\in\mathcal{Y}} d_{W^t}(\bm x_t,\bm \mu^t_y).   
\end{align}      

The threshold parameter $\tau$ is used to determine that an instance does not belong to one of the known classes.
We assign an instance $\bm x_t$ to the \texttt{unknown} class if the confidence of the nearest class $C_{y*} \le \tau^{t}$.
We also learn $\tau^t$ incrementally from the data, as the mean of the confidence values $C_y$ observed since the last added novel class, and it is given by:
\begin{equation}
	\tau^{t+1} = \begin{cases}
        0, &\textrm{if $y_t$ is from a novel class}  \\    
    	 (1-\tfrac{1}{t^*}) \ \tau^{t} + \frac{1}{t^*} C_{y_t}(\bx_t,\theta^t), & \textrm{otherwise}
     \end{cases}
     \label{eq:tau_est}
\end{equation}
where $\bx_t$ is the current training sample and $t^*$ is the number of training samples  since the last addition of a novel class. The value of $\tau$ can be seen as the expected value of the internal confidence associated with the observed training data.

For learning the means and the metric $W$, we resort to the incremental NCM updates defined in \Eqs{inc_ncm_m}{inc_ncm_w}.
A known limitation of the class mean models is the limited flexibility of the representation, which results in linear classifiers.
In the next section we introduce a local learning approach which allows for non-linear classification. 


\subsection{Local Learning in the Open World}
To achieve non-linearity through local learning, 
we use a nearest ball classifier, see~\fig{nno-illustration} (\emph{right}), where balls are added incrementally and combine it with incremental metric learning. 
A ball is defined by its center $\bm c_b \in \mathbb{R}^d$ and its radius $\epsilon_b$. It has a local class probability $p_b(y) = \frac{n_b(y)}{n_b}$, where $n_b(y)$ is the number of (training) samples within this ball assigned to class y and $n_b$ is the total number of samples assigned to this ball. 
For predicting the class label of an example $\bm x$, the ball classifier uses the local class probability $p_{b^*}$ for the nearest ball $b^* = \argmin_{b\in\sS} \ d_{W}(\bm x, \bm c_b)$, where $\sS$ is the current set of covering balls.
To learn the set of balls we follow~\cite{derosa2015abacoc}, which uses the $\ell_2$ distance (\ie $m=d$, and the identity matrix for $W$).
During training, the sequence of observed training examples is used to incrementally build a set $\sS$ of balls that cover the region of the feature space they span. 
At time step $t$, let $b^*$
denote the nearest ball of training example $\bm x_t$, then the updates are:
\begin{description}
	\item[\textbf{if} $d_{W}(\bm x_t,\bm c_{b^*}) > \epsilon_{b^*}$]\mbox{}\\
   The example falls beyond the nearest ball and is used to create a new ball $b'$ which is added to the current set $\sS$ of balls.
   This ball is initialized with:
   \begin{align}
   		\bm c^{t+1}_{b'} &= \bm x_t,\\
		\epsilon^{t+1}_{b'} = \epsilon^0_{b'} &=  d_{W}(\bm x_t,\bm c_{b^*}) , 
   \end{align}   
   the radius is set to the distance to the nearest current ball $b^*$ in order to span the full space between $\bm x_t$ and $\bm c_{b^*}$.
   The label $y_t$ is used to initialize the local class probability $p_{b'}$.    
   \item[\textbf{otherwise}] \mbox{}\\
    The example is considered to belong to the ball $b^*$, and the local class probability is updated using $y_t$.
    The mean and radius are updated depending on the predicted class label $\hat{y}_t = \argmax_y p_{b^*}(y)$:
    \begin{align}
    		\bm c^{t+1}_{b^*} &= \left(1-\frac{1}{n_{b^*}}\right) \bm c^{t}_{b^*} + \frac{1}{n_{b^*}} \bm x_t &&\textrm{if } \hat{y}_t = y_t \textrm{ (correct prediction)},\\
		\epsilon^{t+1}_{b^*} &= \epsilon^0_{b^*} \ m_{b^*}^{-1/(2+\hat{d})}  && \textrm{if } \hat{y}_t \neq y_t \textrm{ (local classifier mistake)},
    \end{align}
  where $\hat{d}$ is the intrinsic dimension of the space (which we fix to $m$ of the low-rank matrix $W$ in the experiments).
	The mean is updated using only correctly predicted samples $\bm x_t$.
    The radius is updated using the initial radius $\epsilon^0_{b^*}$ and $m_{b^*}$ a count of the number of errors made within this ball so far.   
\end{description}

\noindent 
While this training procedure incrementally adds novel balls, it is not designed to predict unknown classes, and it uses the standard $\ell_2$ distance metric.

\myparagraph{Novelty detection} The ball classifier has two important local properties, the local class probability and the ball radius.
The latter could be seen as an indicator of the local complexity in the feature space: if the feature space is locally smooth with respect to the class labels, the radius is likely to be large for this ball, while for a complex, non-smooth feature space the ball radius will be small.
We combine these two properties for the estimation of the prediction confidence.

Given the nearest ball $b^*$ for the example $\bm x_t$, we estimate the prediction confidence as follows:
\begin{align}
C'_y(\bm x_t,b^*) = p_{b^*}\!(y) \ \exp\left( - \frac{1}{2 \epsilon_{b^*}} \ d_W(\bm x_t,\bm c_{b^*}) \right) , \label{eq:abacoc_rbf}
\end{align}
which combines the local class probability $p_{b^*}$, with the RBF kernel estimate, 
where the local bandwidth is set to twice the radius of the ball $\epsilon_{b^*}$.
Intuitively, it assigns the highest confidence to the examples closer to a ball with a pure distribution. 
As opposed to global bandwidth in NNO, we use local bandwidths defined by the ball radii. 

The threshold parameter $\tau$, used to assign instances to the \texttt{unknown} class, is learned incrementally similar to \eq{tau_est}, albeit only using samples which are assigned to ball $b^*$ (\ie $d_{W}(\bm x_t,\bm c_{b^*}) \le \epsilon_{b^*}$), and using the confidence function \eq{abacoc_rbf}.
Since the NBC uses more class centroids (compared to NCM/NNO) the estimate $\tau^t$ converges slowly to the true value. 
To mitigate this problem, we use the Hoeffding bound~\cite{hoeffding1963probability}, since we consider the input samples i.i.d and the confidence \eq{abacoc_rbf} is limited in $[0,1]$, defined as:
\begin{equation} \label{eq:bound_exp_conf}
	\overline{\tau}^t= \tau^t + \sqrt{\frac{1}{2t^*}\cdot\log{\frac{1}{\delta}}}, 
\end{equation}   
where $\delta$ is the desired confidence level, which we set inversely proportional to the time $t$ and number of current classes $C$, $\delta=\frac{1}{t^* C}$.
This bound becomes closer to $\tau^t$ with increasingly more training examples, and less tight when the number of classes increase.
For novelty prediction we assign an instance to the \texttt{unknown} class when $C'(\bm x_t,b^*)<\overline{\tau}^t$. 

\myparagraph{Metric learning} 
For learning the metric $W$, we use a non-linear variant of the NCM classifier. 
We define the class probability of class $y$ as:
\begin{align}
	p_{\textrm{NBC}}(y|\bm x) &= \frac{\sum_{b \in \sS_y} \exp\left(-\tfrac{1}{2} d_{W^t}(\bm x,\bm c_b) \right)}{\sum_{b' \in \sS} \exp\left(-\tfrac{1}{2} d_{W^t}(\bm x,\bm c_{b'}) \right)},
\end{align}
where $\sS_y$ denotes the set of balls which are assigned to class $y$, for this assignment we use a majority vote, \ie $y = \argmax p_b$.
At each time step we do a single SGD update of the metric $W^t$ \wrt the log-likelihood of this model, similar to~\eq{inc_ncm_w}.

This formulation is similar to the non-linear NCM variant proposed in~\cite{ncm_metric}, albeit they used a fixed number of centroids per class and k-means to determine these centroids \emph{a priori}.
In contrast our method learns the number of balls, the number of balls per class and the centroids of each ball incrementally.
\section{Experiments}
In this section we validate our online metric learning approaches on three different validation scenarios.
We show that all three proposed extensions, the online metric learning, the incremental updating of the thresholds, and the local ball classifier lead to better predictions on two different datasets.
We will make available the used features, evaluation protocols and data upon publication.
\vspace{-0.2cm}

\subsection{Datasets}
\vspace{-0.3cm}

\myparagraph{ImageNet ILSVRC'10~\cite{ilsvrc10}}
The first dataset we use is the subset of ImageNet used for the ILSVRC'10 challenge. It contains about 1.2M images for training (with $650 \sim 3000$ images per class), 50K images for validation and 150K images for testing.
For this dataset we use densely sampled SIFT features clustered into $1K$ visual words provided in~\cite{ilsvrc10}.
Though more advanced features are available~\cite{sanchez13ijcv,krizhevsky2012imagenet,szegedy14arxiv}, 
this combination of dataset and features allow for fair comparison to the performance of NCM-Forests~\cite{ristin2015incremental} and the original NNO~\cite{bendale2015towards} methods.

\myparagraph{Places-2~\cite{zhou15places2}}
The second dataset we consider is the recent Places-2 dataset, which contains over 10M images of 400 different scene types. 
The dataset features 5000 to 30,000 training images per class, consistent with real-world frequencies of occurrence.
For this dataset, we use deep learning features by training a GoogLeNet style ConvNet~\cite{szegedy14arxiv} on all 15K ImageNet classes which have more than 200 images using Caffe~\cite{jia2014caffe}.
Subsequently we process the images of the Places-2 dataset and extract the final last 1024 dimensional layer as image representation.

\begin{table}[t]
\centering
	\caption{Comparison on incremental learning on the ILSVRC'10 dataset, all using the same features. The bottom two rows show our proposed incremental metric learning approaches, the other results are taken from~\cite{ristin2015incremental}. The two incremental metric learning algorithms clearly outperform other methods when the number of classes increases}
\vspace{1mm}
\setlength{\tabcolsep}{10pt}
\begin{tabular}{|l c c c c c|}
	\hline
	\textsc{method} \textbackslash \hspace{0.1cm} \textsc{\# of classes}  & \textbf{50}   & \textbf{100}  & \textbf{200}  & \textbf{500}  & \textbf{1000} \\ \hline
	\hline
	\multicolumn{6}{|l|}{\textbf{Baselines} --- results from ~\cite{ristin2015incremental}}\\[1mm]
	Multi-class SVM~\cite{multiclass_svm}        & 42 & 34 & 22 & 10 & 5 \\
	SVM-Forest~\cite{ristin2015incremental}       & \textbf{47} & \textbf{38} & 29 & 19 & 14 \\
	NCM~\cite{ncm_metric}          					& 44 & 36 & 27 & 19 & 14 \\\hline
	\multicolumn{6}{|l|}{\textbf{Incremental learning}  --- results from ~\cite{ristin2015incremental}}\\[1mm]
	NCM-Fix metric                                               				& 32 & -& - & 9 & 6\\
	NCM-Forest													& 41 & - & - & 16 & 11\\
	SVM-Forest													&45 & - & - & 19 & 14\\\hline
	\multicolumn{6}{|l|}{\textbf{Online learning --- this paper}}\\[1mm]
	\textbf{oNCM}                                                			& 42 & 37 & \textbf{32} & \textbf{24} & \textbf{19} \\
	\textbf{oNBC}                                       					& 42 & 34 & 30 & 21 & 16 \\
	\hline
\end{tabular}
\label{tab:rustin}	
\end{table}

\subsection{Scenario 1: Large-Scale Incremental Learning}
In this experiment we follow a large-scale incremental learning scenario as used by~\cite{ristin2015incremental}.
The experimental setup is as follows:
\begin{itemize}
	\item Parameters and metric (if relevant) are learned on an initial set of 20 classes;
	\item Classes are incrementally added in batches of $10$ classes;
	\item Performance is evaluated on the test set after $100, 500$, and $1000$ classes.
\end{itemize}
We use the best performing incremental methods from~\cite{ristin2015incremental} for comparison, specifically: NCM with initial metric, NCM-Forest, and SVM-Forest.
We compare against three non-incremental baselines: multi-class SVMs~\cite{multiclass_svm}, metric learning NCM~\cite{ncm_metric}, and SVM-Forest~\cite{ristin2015incremental}.
We use our online oNCM and oNBC (without novelty detection) in this comparison. 
Our methods are learned incrementally from the start, while shuffling the data within each batch before learning.
For the whitening of the features (to avoid numerical instabilities), we use the mean and standard deviation calculated on the initial set of 20 classes.
Performance is measured using the Top-1 Accuracy, as  commonly used on the ILSVRC dataset.

Results are shown in \tab{rustin}; we highlight two findings. 
First, we observe that among metric learning approaches, the NCM variants are on par with SVM approaches. 
Second, 
we notice that the performance for all algorithms decreases as the number of classes increases. This is to be expected, as the classification problem becomes harder as the number of classes grows. Still, the decrease is definitely more graceful when the metric is being learned incrementally, as for oNCM and oNBC.
We believe this is mainly due to the incremental learning of the metric that leads to continuously adapting to the new classes, rather than relying only on the initial, limited knowledge of the problem. 



\begin{figure}[t]
	\centering
  	\includegraphics[width=.9\linewidth,height=4cm]{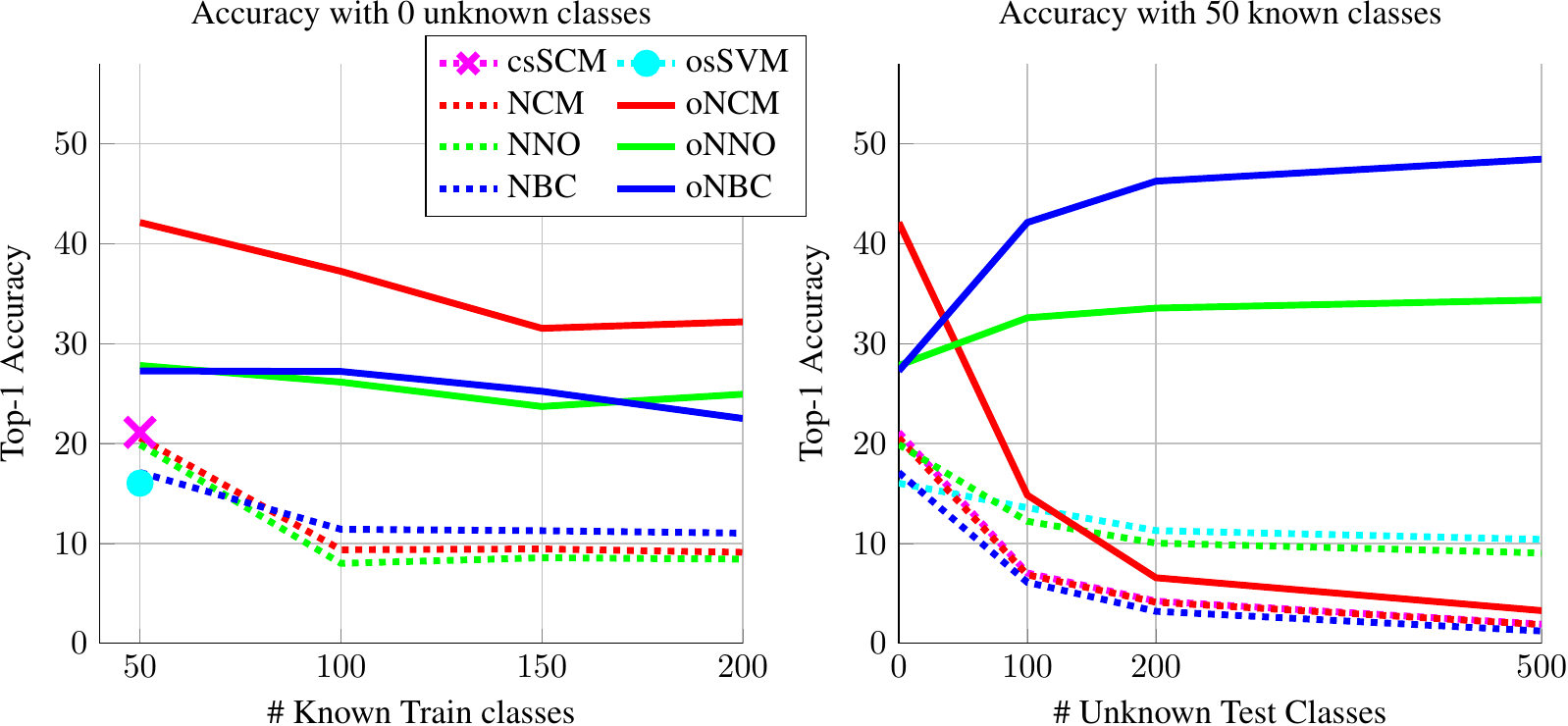}
	\caption{Comparison of results on the open world recognition on the ILSVRC'10 dataset. The proposed incremental/online algorithms oNCM, oNNO, and oNBC clearly outperform their non incremental counterparts.}
	\vspace{-0.5cm}
    \label{fig:openworld}
\end{figure}

\begin{figure}[b]
	\centering
	\begin{minipage}[b]{.45\textwidth}
	\includegraphics[width=\linewidth,height=4cm]{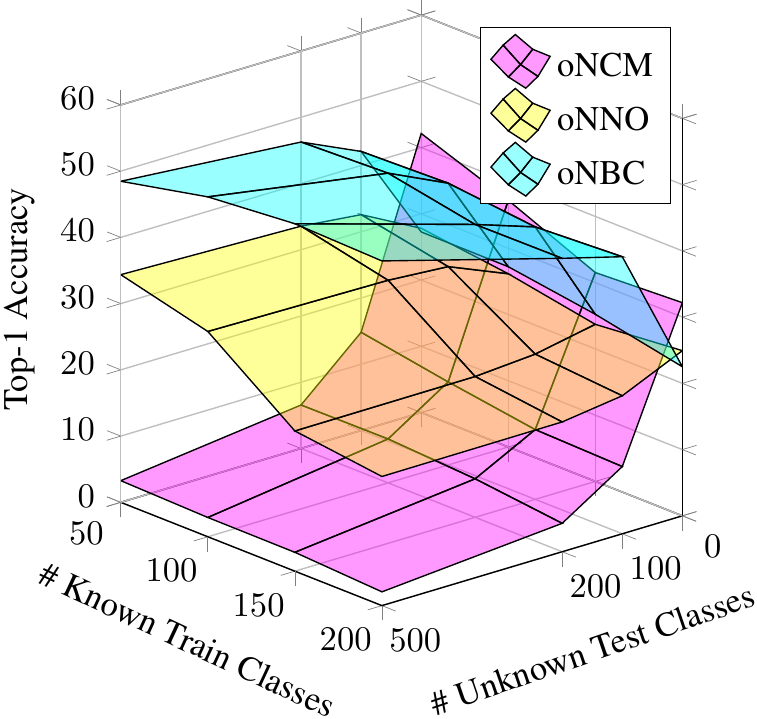}
	\end{minipage}\quad%
  	\begin{minipage}[b]{.3\textwidth}
		\caption{\small{Surface plot of the proposed open world online metric learning methods. The local learning oNBC method clearly outperforms the other method when the number of unknown test classes increases.}}		
		\vspace{-0.5cm}
        \label{fig:openworld2}
	\end{minipage}
\end{figure}

\subsection{Scenario 2: Open World Recognition}
In this experiment we follow the open world protocol proposed in~\cite{bendale2015towards}, where methods are tested on both known and unknown classes.
The experimental setup is as follows:
\begin{itemize}
	\item Parameters and metric are learned on an initial set of 50 classes;
	\item Images of 50 classes are added in each iteration;
	\item Performance is evaluated on a test-set of known  and unknown classes.
\end{itemize}
The open world performance is measured considering  the unknown classes as a single new category.
This allows us to calculate the standard multi-class 1-top accuracy $\hat{y} = \argmax_{y \cup \texttt{unknown}} s_y(\bm x)$, as in~\cite{bendale2015towards}.

We compare our proposed methods against several baselines.
First, we evaluate against a standard linear SVM~\cite{fan2008liblinear} and the 1vSet SVM~\cite{scheirer2013toward}. 
The latter is designed for open-set recognition, which allow to classify images of unknown classes; note that this method is not able to learn incrementally new classes.
We also compare against  NCM~\cite{ncm_metric}, NNO~\cite{bendale2015towards} and NBC~\cite{derosa2015abacoc}, which all allow to adjust towards new classes in an incremental way.
Of these three methods, only NNO is designed to assign images to an unknown class. 
NCM and NNO train their metric on the initial set, and NBC is using the $\ell_2$ metric with the incremental ball set construction.

We use our online oNCM, oNNO, and oNBC in this comparison, all trained incrementally from the start.
Both oNNO and oNBC are able to assign images to unknown classes, while oNCM does not have this property. 

To assess performance in the open world recognition setting one had to consider 
two variables: the number of known categories in incremental learning, and the number of unknown categories during testing. 
We visualize our results in~\fig{openworld}. On the left, we show the top-1 accuracy as the number of known training classes grows, in the case of 0 unknown classes. On the right, we show how the top-1 accuracy changes as the number of unknown test classes increases, for a fixed number of known classes (set to 50).   

Our main observation is that our online approaches clearly outperform all the other in both the closed set and open world settings. 
The lack of rejecting images from unknown classes yield the almost random performance of the NCM method.
Note that oNCB adapts to the classification problems and reject images from unknown classes, indeed prediction becomes easier when the number of unknown and known classes are unbalanced.
In ~\fig{openworld2}, we show a surface plot over a different range of known classes and unknown classes for our proposed online methods.

\begin{figure}[t]
	\centering
	\includegraphics[width=.9\textwidth,height=8cm]{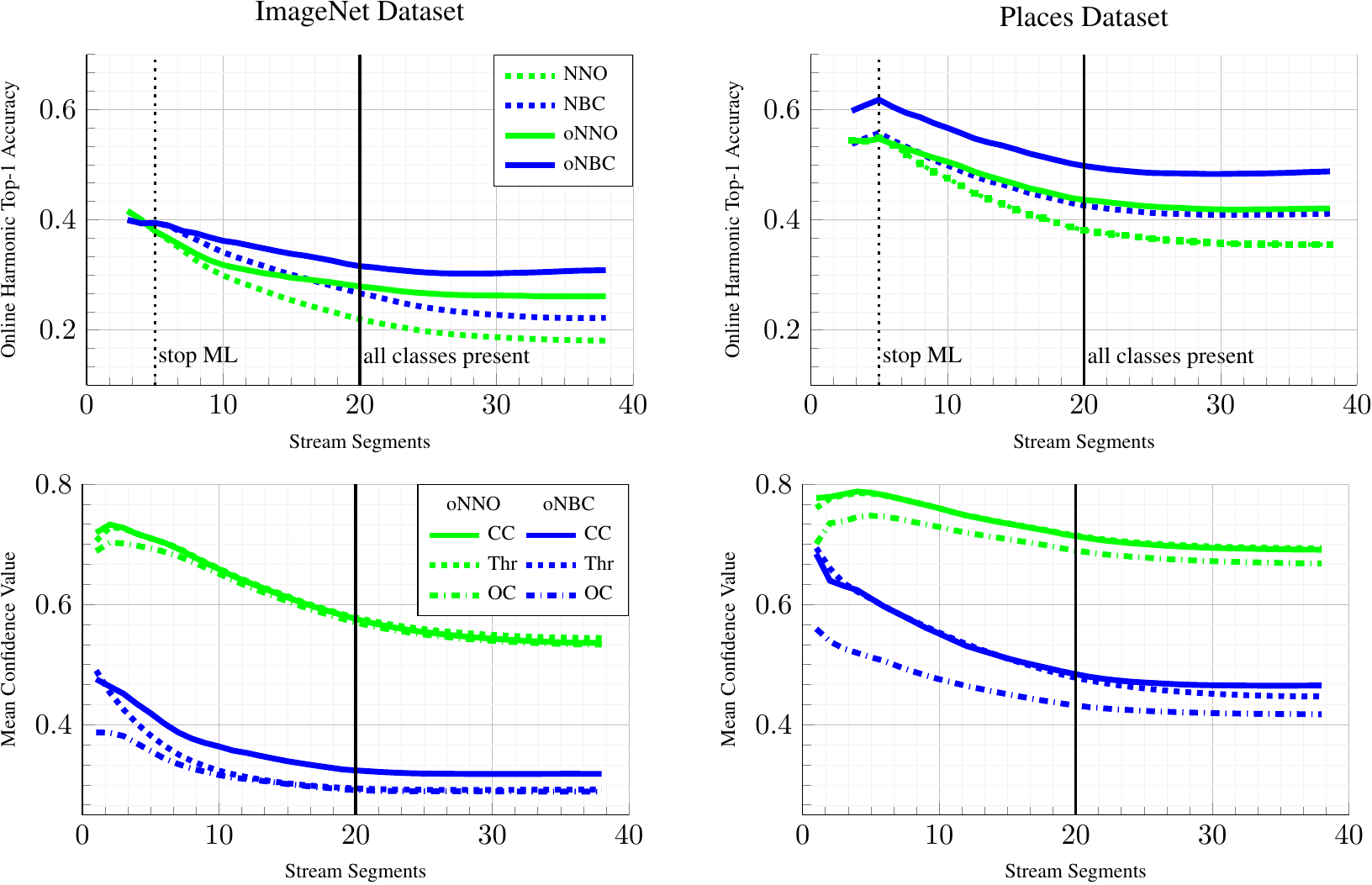}
	\caption{Results on ImageNet and Places-2 dataset}
	\vspace{-0.5cm}
    \label{fig:51_ip}
\end{figure}
\subsection{Scenario 3: Online Image Stream Prediction}
In this experiment, we aim to simulate an online image stream prediction setting, for which we introduce a novel evaluation protocol.
We believe it is a more realistic protocol, that permits to fully represents the dynamical behavior of the algorithm simultaneously during the updating and testing phases. 
The experimental setup we consider follows~\alg{openworld}, where we consider a stream of incoming images. At time $t$ the learner:
\begin{enumerate}
    	\item Predicts the label $\hat{y}_t$ for sample $\bm x_t$ using the current models;
  	\item Updates the online accuracy using $\hat{y}_t$ and the ground-truth label $y_t$;
  	\item Updates the current models using training tuple $(\bm x_t,y_t)$.
\end{enumerate}
	
For practical reasons we generate the stream from 1200 images of each of the 200 most frequent classes from ILVRC'10 and Places-2, 100 classes as being known and 100 for the unknown classes. In this way the final number of instances for both close an open set classes is totally balanced.
The data stream is generated as follows:
\begin{enumerate}
	\item The stream is divided in 40 stream-segments;
	\item The first 20 segments introduce 5 known and 5 unknown classes each;				
	\item The learner is given 60 images per active class per segment;
	\item Any introduced class dries up after 20 segments;
	\item The number of images per segment varies, with a peak half-way;
	\item The online accuracy is recorded after each of the 40 stream-segments.
\end{enumerate}	
We believe this setting is interesting because the evaluated known and unknown classes evolve over time, both by increasing the number of classes as well as reducing the number of classes.

For evaluating the performance of the stream, we use the online accuracy~\cite{gama2013evaluating}  of the harmonic mean ($\mathrm{hm}^t$, also known as the F-Score) between the closed set accuracy and open set accuracy as follows:
\begin{align*}
	 \mathrm{hm}^t  &= 2 \frac{A^{t^o}_{\textrm{o}} \cdot A^{t^c}_{\textrm{c}}}{A^{t^o}_{\textrm{o}} + {A^{t^c}_\textrm{c}}},& \mbox{where } 	A^{t^c}_{\textrm{c}}= &\bigl(1-\tfrac{1}{t^c}\bigr)A_{t^c-1} + \tfrac{1}{t^c}\ind{\yhat_{t^c} = y_{t^c}},\\ 	& 	& \mbox{ and }A^{t^o}_{\textrm{o}}=&\bigl(1-\tfrac{1}{t^o}\bigr)A_{t^o-1} + \tfrac{1}{t^o}\ind{\yhat_{t^o} = \texttt{unknown}},
\end{align*}
for $t^c \in t \mid y_{t}\in\texttt{known}$ and $t^o  \in t \mid y_{t}\in\texttt{unknown}$.
We coin this method the \emph{online harmonic top-1 accuracy}.
This equally weights the performance of closed set accuracy and open set accuracy. 
Moreover, a method which performs well on one of the two accuracies and poorly on the other obtains a low harmonic mean, which is a favorable property.

For this experiment we use the NNO and NBC methods  on the ILVRC'10 and Places2 dataset.
The results of this experiment are presented in~\fig{51_ip}, in which we compare oNNO and oNBC to variants using just an initial learned metric, these are learned in an online learning phase of 5 stream-segments (and indicated by NNO/NBC in the figure).
In the top-row figures, we show the online harmonic accuracy, and once again the incremental metric learning methods oNNO and oNBC have a clear benefit over their (fixed) metric-learning counterparts. This becomes clearer when more images and classes are added in later stream segments.
Moreover, the local learning NBC classifier can adjust more precisely to the added classes and therefore outperforms the linear NNO classifier. Notice that after no addition of new classes both methods start to gain performances as they are learning the already explored categories.
Finally, the significant difference in performance between the ILVRC'10 and Places2 datasets, while using the same amount of classes and images, is likely to be due to the more powerful features used for the Places2 dataset. %

In the bottom row figures of~\fig{51_ip}, we show the mean of the confidence values assigned to the  closed set (CC)  and the open set (OC), together with the mean of the estimated thresholds (Thr) by our methods within each stream-segment.
In order to achieve good performances (open and close), the threshold for rejecting an image into the \texttt{unknown} class should lie between the closed set and open set confidence. 
From the results, it can be observed that the open set and closed set confidence is almost identical for the oNNO classifier, therefore finding a good threshold value is almost impossible.
For the oNBC method, where the confidence function and 
the estimated threshold depend on more local information, the open set and closed set confidences are well set apart. We remark that using a fixed threshold tuned on an initial set (as the literature methods do) can not lead to good performances as the confidence change over time.
\vspace{-0.3cm}

\section{Conclusions}
\vspace{-0.3cm}

In this paper we addressed the open world recognition problem and proposed three extensions to its current formulation: online metric learning, incremental updating of thresholds for novelty detection, and local learning through nearest ball classification. We evaluated the effect of these extensions over three different existing algorithms, NCM, NNO and NBC, and we assessed the effects of our extensions over three different experimental scenarios: large-scale incremental learning, open world recognition and online image stream prediction. This last setting is a new protocol for evaluation of online open world recognition, which we believe mimics better out-of-the-lab applications. For all the three scenarios, our proposed methods performed substantially better than the baselines, showcasing the importance of fully embracing online learning for open world recognition.

Future work will focus on studying the suitability of active learning in this scenario~\cite{joshi2009multi}, where an  interaction module has to balance the number of true label requests and the performance at any query rate. Another setting we will investigate will be the bandit one~\cite{bubeck2012regret} where the learners can access to the labels only when they are making correct predictions. 




{\small

\begin{thebibliography}{10}
 	
 	\bibitem{multiclass_svm}
 	Z.~Akata, F.~Perronnin, Z.~Harchaoui, and C.~Schmid.
 	\newblock Good practice in large-scale learning for image classification.
 	\newblock {\em IEEE Trans.\ PAMI}, 2013.
 	
 	\bibitem{bendale2015towards}
 	A.~Bendale and T.~Boult.
 	\newblock Towards open world recognition.
 	\newblock In {\em CVPR}, 2015.
 	
 	\bibitem{ilsvrc10}
 	A.~Berg, J.~Deng, and L.~Fei-Fei.
 	\newblock The {ImageNet} large scale visual recognition challenge 2010-2015.
 	\newblock http://www.image-net.org/challenges/LSVRC/2015, 2010.
 	
 	\bibitem{bubeck2012regret}
 	S.~Bubeck and N.~Cesa-Bianchi.
 	\newblock Regret analysis of stochastic and nonstochastic multi-armed bandit
 	problems.
 	\newblock {\em arXiv preprint arXiv:1204.5721}, 2012.
 	
 	\bibitem{crammer2006online}
 	K.~Crammer, O.~Dekel, J.~Keshet, S.~Shalev-Shwartz, and Y.~Singer.
 	\newblock Online passive-aggressive algorithms.
 	\newblock {\em Journal of Machine Learning Research}, 7:551--585, 2006.
 	
 	\bibitem{de2014online}
 	R.~De~Rosa, N.~Cesa-Bianchi, I.~Gori, and F.~Cuzzolin.
 	\newblock Online action recognition via nonparametric incremental learning.
 	\newblock In {\em BMVC}, 2014.
 	
 	\bibitem{derosa2015abacoc}
 	R.~De~Rosa, F.~Orabona, and N.~Cesa-Bianchi.
 	\newblock The {ABACOC} algorithm: a novel approach for nonparametric
 	classification of data streams.
 	\newblock In {\em ICDM}, 2015.
 	
 	\bibitem{deng2011fast}
 	J.~Deng, S.~Satheesh, A.~C Berg, and F.~Li.
 	\newblock Fast and balanced: Efficient label tree learning for large scale
 	object recognition.
 	\newblock In {\em NIPS}, 2011.
 	
 	\bibitem{everingham2010pascal}
 	M.~Everingham, L.~Van~Gool, C.~KI Williams, John Winn, and A.~Zisserman.
 	\newblock The pascal visual object classes (voc) challenge.
 	\newblock {\em IJCV}, 88(2):303--338, 2010.
 	
 	\bibitem{fan2008liblinear}
 	R.~Fan, K.~Chang, C.~Hsieh, X.~Wang, and C.~Lin.
 	\newblock Liblinear: A library for large linear classification.
 	\newblock {\em Journal of Machine Learning Research}, 9:1871--1874, 2008.
 	
 	\bibitem{fragoso2013evsac}
 	V.~Fragoso, P.~Sen, S.~Rodriguez, and M.~Turk.
 	\newblock Evsac: accelerating hypotheses generation by modeling matching scores
 	with extreme value theory.
 	\newblock In {\em ICCV}, 2013.
 	
 	\bibitem{gama2013evaluating}
 	J.~Gama, R.~Sebastiao, and P.~Rodrigues.
 	\newblock On evaluating stream learning algorithms.
 	\newblock {\em Machine Learning}, 2013.
 	
 	\bibitem{hoeffding1963probability}
 	W.~Hoeffding.
 	\newblock Probability inequalities for sums of bounded random variables.
 	\newblock {\em Journal of the American statistical association},
 	58(301):13--30, 1963.
 	
 	\bibitem{jain2014multi}
 	L.~P Jain, W.~J Scheirer, and T.~E Boult.
 	\newblock Multi-class open set recognition using probability of inclusion.
 	\newblock In {\em ECCV}, 2014.
 	
 	\bibitem{jia2014caffe}
 	Yangqing Jia, Evan Shelhamer, Jeff Donahue, Sergey Karayev, Jonathan Long, Ross
 	Girshick, Sergio Guadarrama, and Trevor Darrell.
 	\newblock Caffe: Convolutional architecture for fast feature embedding.
 	\newblock {\em arXiv preprint arXiv:1408.5093}, 2014.
 	
 	\bibitem{joshi2009multi}
 	A.~J Joshi, F.~Porikli, and N.~Papanikolopoulos.
 	\newblock Multi-class active learning for image classification.
 	\newblock In {\em CVPR}. IEEE, 2009.
 	
 	\bibitem{krizhevsky2012imagenet}
 	A.~Krizhevsky, I.~Sutskever, and G.~E Hinton.
 	\newblock Imagenet classification with deep convolutional neural networks.
 	\newblock In {\em NIPS}, 2012.
 	
 	\bibitem{Kuzborskij_2013_CVPR}
 	Ilja Kuzborskij, Francesco Orabona, and Barbara Caputo.
 	\newblock From n to n+1: Multiclass transfer incremental learning.
 	\newblock In {\em The IEEE Conference on Computer Vision and Pattern
 		Recognition (CVPR)}, June 2013.
 	
 	\bibitem{laskov2006incremental}
 	P.~Laskov, C.~Gehl, S.~Kr{\"u}ger, and K.~M{\"u}ller.
 	\newblock Incremental support vector learning: Analysis, implementation and
 	applications.
 	\newblock {\em Journal of Machine Learning Research}, 7:1909--1936, 2006.
 	
 	\bibitem{li2005open}
 	F.~Li and H.~Wechsler.
 	\newblock Open set face recognition using transduction.
 	\newblock {\em IEEE Trans.\ PAMI}, 27(11):1686--1697, 2005.
 	
 	\bibitem{li2010optimol}
 	L.~Li and L.~Fei-Fei.
 	\newblock Optimol: automatic online picture collection via incremental model
 	learning.
 	\newblock {\em IJCV}, 88(2):147--168, 2010.
 	
 	\bibitem{liu2013probabilistic}
 	B.~Liu, M.~Sadeghi, F.and~Tappen, O.~Shamir, and C.~Liu.
 	\newblock Probabilistic label trees for efficient large scale image
 	classification.
 	\newblock In {\em CVPR}, 2013.
 	
 	\bibitem{marszalek2008constructing}
 	M.~Marsza{\l}ek and C.~Schmid.
 	\newblock Constructing category hierarchies for visual recognition.
 	\newblock In {\em ECCV}, 2008.
 	
 	\bibitem{ncm_metric}
 	T.~Mensink, J.~Verbeek, F.~Perronnin, and G.~Csurka.
 	\newblock Distance-based image classification: Generalizing to new classes at
 	near-zero cost.
 	\newblock {\em IEEE Trans.\ PAMI}, 35(11):2624--2637, 2013.
 	
 	\bibitem{poggio2001incremental}
 	T.~Poggio.
 	\newblock Incremental and decremental support vector machine learning.
 	\newblock In {\em NIPS}, 2001.
 	
 	\bibitem{pronobis2006more}
 	B.~Pronobis, A.and~Caputo.
 	\newblock The more you learn, the less you store:
 	memory$\backslash$--controlled incremental svm.
 	\newblock Technical report, IDIAP, 2006.
 	
 	\bibitem{ristin2015incremental}
 	M.~Ristin, M.~Guillaumin, J.~Gall, and L.~van Gool.
 	\newblock Incremental learning of ncm forests for large-scale image
 	classification.
 	\newblock In {\em IEEE Trans.\ PAMI}, 2016.
 	
 	\bibitem{sanchez13ijcv}
 	J.~S\'{a}nchez, F.~Perronnin, T.~Mensink, and J.~Verbeek.
 	\newblock Image classification with the {F}isher vector: Theory and practice.
 	\newblock {\em IJCV}, 2013.
 	
 	\bibitem{scheirer2013toward}
 	W.~J Scheirer, A.~de~Rezende~Rocha, A.~Sapkota, and T.~E. Boult.
 	\newblock Toward open set recognition.
 	\newblock {\em IEEE Trans.\ PAMI}, 35(7):1757--1772, 2013.
 	
 	\bibitem{scheirer2014probability}
 	W.~J Scheirer, L.~P Jain, and T.~E Boult.
 	\newblock Probability models for open set recognition.
 	\newblock {\em IEEE Trans.\ PAMI}, 36(11):2317--2324, 2014.
 	
 	\bibitem{shalev2011online}
 	S.~Shalev-Shwartz.
 	\newblock Online learning and online convex optimization.
 	\newblock {\em Foundations and Trends in Machine Learning}, 4(2), 2011.
 	
 	\bibitem{shalev2011pegasos}
 	S.~Shalev-Shwartz, Y.~Singer, N.~Srebro, and A.~Cotter.
 	\newblock Pegasos: Primal estimated sub-gradient solver for svm.
 	\newblock {\em Mathematical programming}, 127(1):3--30, 2011.
 	
 	\bibitem{simonyan2014very}
 	K.~Simonyan and A.~Zisserman.
 	\newblock Very deep convolutional networks for large-scale image recognition.
 	\newblock Technical report, arXiv preprint arXiv:1409.1556, 2014.
 	
 	\bibitem{szegedy14arxiv}
 	Christian Szegedy, Wei Liu, Yangqing Jia, Pierre Sermanet, Scott Reed, Dragomir
 	Anguelov, Dumitru Erhan, Vincent Vanhoucke, and Andrew Rabinovich.
 	\newblock Going deeper with convolutions.
 	\newblock Technical report, arXiv Preprint arxiv:1409.4842, 2014.
 	
 	\bibitem{veenman2005nearest}
 	C.~Veenman and M.~Reinders.
 	\newblock The nearest subclass classifier: A compromise between the nearest
 	mean and nearest neighbor classifier.
 	\newblock {\em IEEE Trans.\ PAMI}, 27(9):1417--1429, 2005.
 	
 	\bibitem{veenman2005weighted}
 	C.~Veenman and D.~Tax.
 	\newblock A weighted nearest mean classifier for sparse subspaces.
 	\newblock In {\em CVPR}, 2005.
 	
 	\bibitem{wang2010multi}
 	Z.~Wang, K.~Crammer, and S.~Vucetic.
 	\newblock Multi-class pegasos on a budget.
 	\newblock In {\em ICML}, 2010.
 	
 	\bibitem{webb02spr}
 	A.~R. Webb.
 	\newblock {\em Statistical pattern recognition}.
 	\newblock {Wiley}, {New-York, NY, USA}, 2002.
 	
 	\bibitem{yeh2008dynamic}
 	T.~Yeh and T.~Darrell.
 	\newblock Dynamic visual category learning.
 	\newblock In {\em CVPR}, 2008.
 	
 	\bibitem{zhou15places2}
 	B.~Zhou, A.~Khosla, A.~Lapedriza, A.~Torralba, and A.~Oliva.
 	\newblock {Places2}: A large-scale database for scene understanding.
 	\newblock Technical report, ArXiV preprint, 2015.
 	
 \end{thebibliography}

}
\appendix
\end{document}